\documentclass{article}

\usepackage{graphicx} 
\usepackage[edges]{forest}
\usepackage{biblatex} 
\usepackage{hyperref}
\usepackage{amsfonts}
\usepackage{amsmath}
\usepackage{booktabs}
\usepackage{siunitx}
\usepackage{enumitem}
\usepackage{threeparttable}
\usepackage[table]{xcolor}
\usepackage{adjustbox}%
\usepackage{subcaption}%
\usepackage{bm}%
\usepackage{authblk}
\usepackage{multirow}
\usepackage{caption}

\usepackage{tcolorbox}

\title{Semantically Conditioned Diffusion Models for Cerebral DSA Synthesis}

\author[1,2,5]{Qiwen Xu}
\author[1,5]{David R\"ugamer}
\author[2,3]{Holger Wenz}
\author[2,4]{Johann Fontana}
\author[2]{Nora Meggyeshazi}
\author[1,5]{Andreas Bender}
\author[2,3]{Máté E. Maros}

\affil[1]{Department of Statistics, LMU Munich, Munich, Germany}
\affil[2]{Department of Biomedical Informatics (DBMI), Medical Faculty Mannheim, 
Heidelberg University, Mannheim, Germany}
\affil[3]{Clinic for Diagnostic and Interventional Neuroradiology, Medical Faculty Mannheim, Heidelberg University, Mannheim, Germany}
\affil[4]{Department of Anesthesiology and Intensive Care Medicine, BG Trauma Center Tuebingen, Tuebingen, Germany}
\affil[5]{Munich Center for Machine Learning (MCML), Munich, Germany} 
\affil[ ]{\textit{Email:}
\texttt{qiwen.xu@stat.uni-muenchen.de}, 
\texttt{david.ruegamer@stat.uni-muenchen.de}, 
\texttt{andreas.bender@stat.uni-muenchen.de}, 
\texttt{mate.maros@medma.uni-heidelberg.de}} 

\date{February 2026}
\begin{document}

\maketitle

\abstract{
Digital subtraction angiography (DSA) plays a central role in the diagnosis and treatment of cerebrovascular disease, yet its invasive nature and high acquisition cost severely limit large-scale data collection and public data sharing. Therefore, we developed a semantically conditioned latent diffusion model (LDM) that synthesizes arterial-phase cerebral DSA frames under explicit control of anatomical circulation (anterior vs.\ posterior) and canonical C-arm positions. We curated a large single-centre DSA dataset of 99,349 frames and trained a conditional LDM using text embeddings that encoded anatomy and acquisition geometry. To assess clinical realism, four medical experts, including two neuroradiologists, one neurosurgeon, and one internal medicine expert, systematically rated 400 synthetic DSA images using a 5-grade Likert scale for evaluating proximal large, medium, and small peripheral vessels. The generated images achieved image-wise overall Likert scores ranging from 3.1 to 3.3, with high inter-rater reliability (ICC(2,\emph{k}) = 0.80--0.87). Distributional similarity to real DSA frames was supported by a low median Fréchet inception distance (FID) of 15.27. Our results indicate that semantically controlled LDMs can produce realistic synthetic DSAs suitable for downstream algorithm development, research, and training.
}


\section{Introduction}


Digital subtraction angiography (DSA) is widely used for visualizing cerebrovascular anatomy, guiding diagnosis and treatment of neurovascular disease~\cite{singh2023digital}. Despite its diagnostic and therapeutic value, DSA acquisition is invasive, involves ionizing radiation and the intraarterial application of iodinated contrast agents, hence it is difficult to perform at a population scale~\cite{singh2023digital}. Publicly accessible DSA datasets remain scarce due to privacy constraints and the effort required to generate, curate and annotate dynamic vascular sequences~\cite{liuDIASDatasetBenchmark2024,Xu2026,zhang2025dsca}. These limitations restrict the training of data-intensive models for segmentation, anomaly detection, or image enhancement, and motivate generative approaches that can provide realistic virtual angiograms for augmentation, simulation, and method development~\cite{kim2023Diffusion,ovalle-magallanes2022Improving}.


Most existing work on generative modeling for DSA or X-ray angiography has focused on generative adversarial network (GAN)-based architectures and task-driven formulations. Examples include generating pseudo-angiograms to support segmentation or registration~\cite{hamdi2024CASGAN,ovalle-magallanes2022Improving,yu2020AFSEG}, domain translation between native angiograms and angiographic views~\cite{gao2019Deep,Liu2024} or synthesizing future frames conditioned on past sequences~\cite{azizmohammadi2022Generative,kim2023Diffusion}. These methods typically operate at limited resolution, rely on an observed input image, and are not designed to generate large, diverse libraries of stand-alone DSA frames with explicit semantic control over anatomical territory or a view depending on the C-arm positioning. Therefore, systematic high-fidelity synthesis of DSA images ``from scratch'' remains largely unexplored.


Diffusion models~\cite{ho2020denoising}, and in particular latent diffusion models (LDMs)~\cite{rombach2022high}, have recently achieved state-of-the-art performance in high-resolution image synthesis and controllable generation in natural-image domains. Extensions to the medical imaging domain have demonstrated promising results for MRI anomaly detection~\cite{pinaya2022fast}, tumor segmentation~\cite{wolleb2022diffusion}, and CT image generation~\cite{weber2023cascaded}, but their application to angiography has so far been restricted to small datasets or low-resolution settings, often without semantic control or rigorous clinical evaluation.

In this work, we present a semantically conditioned LDM for the synthesis of cerebral arterial-phase DSA images under explicit control of anatomical circulation, and acquisition geometry. Leveraging a large single-centre clinical dataset, our approach enables the generation of synthetic angiograms guided by clinically meaningful semantic descriptors of vascular territory and C-arm positionings. We evaluate the realism and clinical plausibility of the generated images using an expert reader study complemented by quantitative distributional analyses. We demonstrate the potential of diffusion-based generative models to provide controllable and clinically relevant synthetic DSA data for downstream research, algorithm development, and training.



\section{Results}

\subsection{Study cohort}
The cohort contained 1104 unique DSA studies from 859 patients (351 females, 41.15\%) performed 
at the Clinic for Diagnostic and Interventional Neuroradiology, University Medical Center Mannheim (UMM) between 2015-2021, with an average of 1.29 studies per patient, where each study corresponded to a single clinical examination and may included multiple DSA acquisitions of different anatomical targets (i.e.\ circulations). The mean patient age was 56 years (SD=15.32), ranging from 18 to 91 years with no significant age difference between male and female patients ($t = -0.03$, $p = 0.97$).

All examinations were acquired using a Siemens Healthineers biplane angiography system (Artis~Q), which served as the standard angiographic suite during the study period.



In the subsequent preprocessing stages (Sec.~\ref{Sec: Data Preprocessing}), the same dataset was further analyzed at the series and frame levels, where individual DSA sequences were filtered and decomposed into arterial-phase image frames for generative modeling.

\subsection{Cohort selection and Data preprocessing}

\label{Sec: Data Preprocessing}
\begin{figure}[tbh]
\centering
\includegraphics[width=0.9\textwidth]{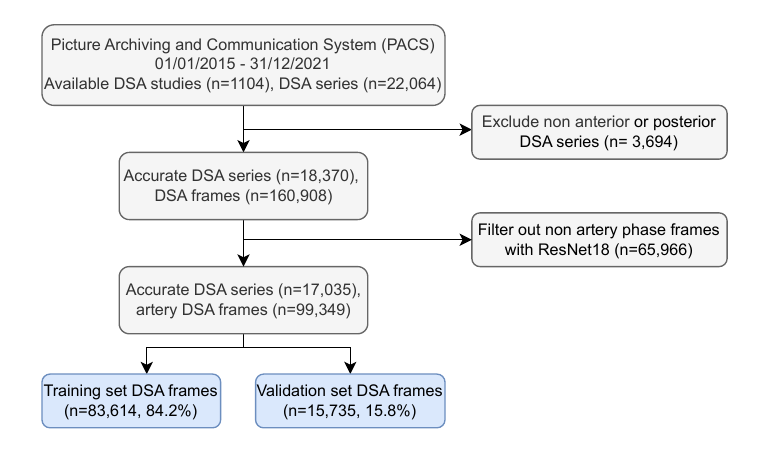}
\caption{Data filtering and preprocessing workflow for constructing the arterial-phase DSA dataset}
\label{fig:data_flow}
\end{figure} 

We applied a multi-stage preprocessing pipeline to obtain arterial-phase DSA frames suitable for training and evaluation (Fig.~\ref{fig:data_flow}), as this phase provides the most complete and clinically relevant visualization of the arterial vasculature and is the primary basis for diagnostic assessment in neuroangiography. Out of a total of 22,064 DSA image series, we excluded studies that did not correspond to anterior or posterior circulations, resulting in the removal of 3694 series (16.7\%), leaving 18,370 valid DSA series for subsequent processing.

To further isolate suitable arterial-phase frames, we trained a ResNet-18 classifier using a subset of the dataset that included 953 out of 18,370 series (5.2\%)—each annotated by an expert neuroradiologist (M.E.M.) based on a single representative arterial-phase frame. Subsequently, on a hold-out validation set, the trained ResNet-18 classifier achieved an arterial-phase detection precision of 0.92 and a recall of 0.89, with an overall validation accuracy of 0.95. The classifier was then applied to the remaining series (N=17,417) to identify arterial-phase frames.
As a result, we obtained a total of 99,349 frames identified as arterial-phase DSA, comprising 78,253 frames (78.77\%) from the anterior circulation (AC) and 21,096 frames (21.23\%) from the posterior circulation (PC).
Within the arterial-phase dataset, we summarized the distribution of semantic conditioning variables used for conditional generation. In particular, we analyzed the frequency of circulation--plane combinations defined by AC/PC and Plane~A/B, as reported in Table~\ref{tab:cond_dist}. Plane~A and Plane~B refer to the two detectors of the biplane angiography system, corresponding to posteroanterior and lateral projections, respectively.

The four conditions selected for the reader study (Sec.~\ref{Sec:Clinical Validation}) correspond to the most prevalent combinations in the training data: AC and PC imaged in Plane~A or Plane~B. We denote these four circulation--plane combinations as AC~A, PC~A, AC~B, and PC~B in the remainder of the manuscript. They are associated with canonical C-arm orientations of $(0^\circ,0^\circ)$ for Plane~A and $(-90^\circ,0^\circ)$ for Plane~B, where the primary and secondary angles denote the two rotational degrees of freedom of the biplane C-arm. Together, these configurations account for 45.4\% of all arterial-phase frames, supporting their use as representative conditions for clinical evaluation.


\begin{table}[thb]
\centering
\begin{threeparttable}
\begin{tabular}{lcc S[table-format=5.0] S[table-format=2.1]}
\toprule
Circulation & Plane & Prim./Sec. ($^\circ$) & {Count} & {Proportion (\%)} \\
\midrule
AC  & A & $0 / 0$    &  9500 &  9.6 \\
PC & A & $0 / 0$    &   920 &  0.9 \\
AC  & B & $-90 / 0$  & 26720 & 26.9 \\
PC & B & $-90 / 0$  &  7912 &  8.0 \\
\midrule
\multicolumn{3}{l}{Others} & 54297 & 54.6 \\
\midrule
\multicolumn{3}{l}{Total}  & 99349 & 100.0 \\
\bottomrule
\end{tabular}
\begin{tablenotes}[flushleft]
\footnotesize
\item Proportions are computed with respect to all arterial-phase frames ($N=99{,}349$).
Angles are discretized into canonical bins using a tolerance of $\pm 5^\circ$ around $0^\circ$ and $-90^\circ$; all other angle combinations are grouped into ``Others''. Plane~A corresponds to a canonical \(0^\circ/0^\circ\) C-arm orientation and Plane~B to \(-90^\circ/0^\circ\), subsequent references to conditioning settings omit explicit angle notation.

\end{tablenotes}
\end{threeparttable}
\caption{Distribution of circulation-plane-angle combinations in the arterial-phase training set}
\label{tab:cond_dist}
\end{table}

\subsection{Synthetic DSA Data Generation}
\label{Sec:results_gen}
To evaluate the quality and clinical realism of our generated DSA images, we conducted a physician assessment study using synthetic samples generated under four commonly observed conditions in the dataset. Specifically, we defined four textual conditioning prompts, which were passed through the BERT-based text 
encoder to produce semantic conditioning embeddings that guide the LDM during image generation.

\begin{tcolorbox}[
    colback=gray!5,
    colframe=gray!50,
    arc=2mm,
    boxrule=0.5pt,
    left=10pt, right=10pt, top=8pt, bottom=8pt
]
\small
\begin{itemize}[leftmargin=*, nosep]
    \item This is an \textbf{anterior} DSA scan taken in \textbf{Plane A}, with a primary angle of 0$^\circ$ and a secondary angle of 0$^\circ$.
    \item This is a \textbf{posterior} DSA scan taken in \textbf{Plane A}, with a primary angle of 0$^\circ$ and a secondary angle of 0$^\circ$.
    \item This is an \textbf{anterior} DSA scan taken in \textbf{Plane B}, with a primary angle of -90$^\circ$ and a secondary angle of 0$^\circ$.
    \item This is a \textbf{posterior} DSA scan taken in \textbf{Plane B}, with a primary angle of -90$^\circ$ and a secondary angle of 0$^\circ$.
\end{itemize}
\end{tcolorbox}


\begin{figure}[htbp]
    \centering
    \includegraphics[width=0.5\textwidth]{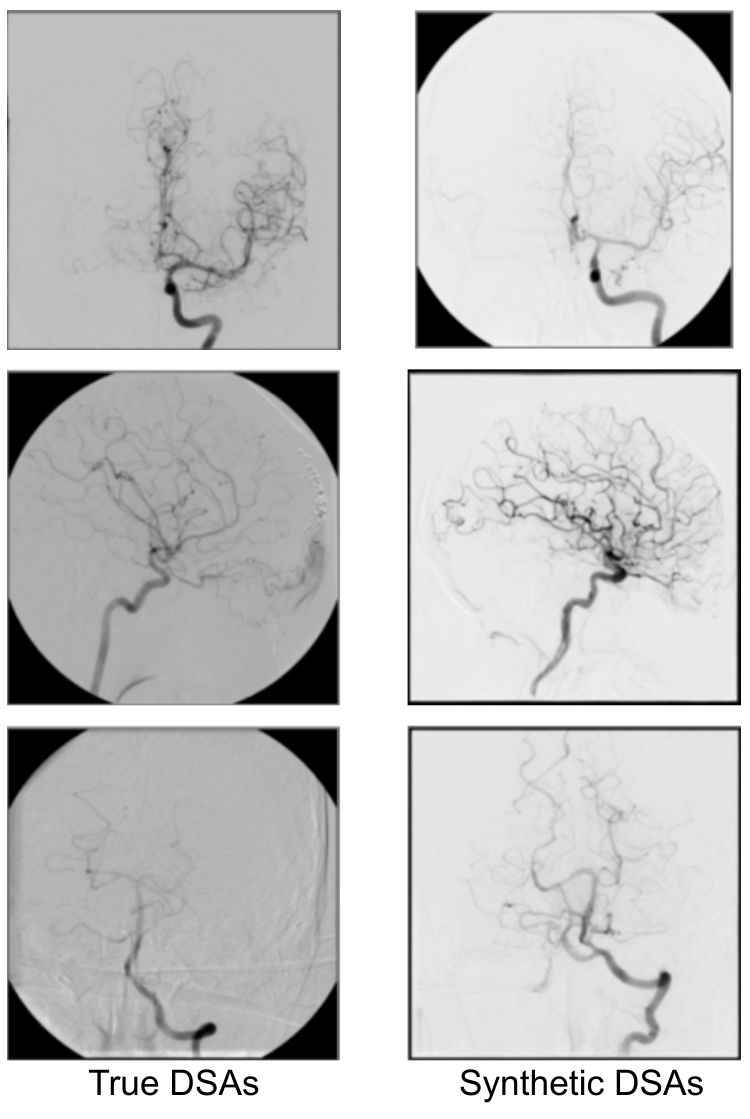}
    \caption{Representative arterial-phase DSA frames from the clinical dataset (left column) and from the conditional diffusion model (right column)}
    \label{fig:dsa_examples}
\end{figure}


These four conditions corresponded to the most frequently occurring anatomical-plane-angle combinations in our curated dataset, ensuring a representative basis for assessment. We generated 10 batches of 40 images sampled under the four conditions above in a stratified manner (10 images per condition per batch), for a total of 400 synthetic DSA frames. This design balances across condition frequencies observed in the real dataset, enabling per-condition analysis with matched sample sizes.
Before conducting the reader study, we visually compared samples from the conditional diffusion model with real arterial-phase DSAs from the same anatomical territories. As illustrated in Fig.~\ref{fig:dsa_examples}, the generated frames closely resemble clinical DSAs in terms of vessel topology and contrast dynamics, providing qualitative support that the model operates in a realistic image regime.

\begin{table}[htbp]
\centering
\resizebox{\textwidth}{!}{
\begin{tabular}{lcccccccc}
\toprule
Circulation & Plane &
\(N_{\text{keep}}\) &
$\bm{N_{\textbf{scored}}}$ &
Prox.\ NA &
Med.\ NA &
Peri.\ NA &
Ext.\ circ. \\
\midrule
AC  & A & 78 & \textbf{70} & 11 &  9 & 12 &  1 \\
AC  & B & 59 & \textbf{46} & 18 & 18 & 19 &  7 \\
PC  & A & 47 & \textbf{46} &  1 &  6 & 13 &  0 \\
PC  & B & 55 & \textbf{55} &  0 &  1 &  3 &  0 \\
\bottomrule
\end{tabular}}

\caption{Screening retention and discrepancies between retained and quality-scored images. $N_{\text{keep}}$ denotes the number of images kept after expert screening; $N_{\text{scored}}$ denotes the number of images with at least one ratable arterial segment. 
Prox./Med./Peri.\ refer to proximal, medium, and peripheral arterial segments, respectively; entries marked as NA count images in which the corresponding segment was labeled as \emph{not present} or \emph{not applicable} by all raters. 
Ext.\ circ.\ counts images consistently labeled as \emph{external circulation}}
\label{tab:screening_overall}
\end{table}

\subsection{Clinical Assessment}
\label{Sec:Clinical Validation}
\begin{figure}[htbp]
\centering
\includegraphics[width=\textwidth, trim=0cm 20cm 0cm 0cm, clip]{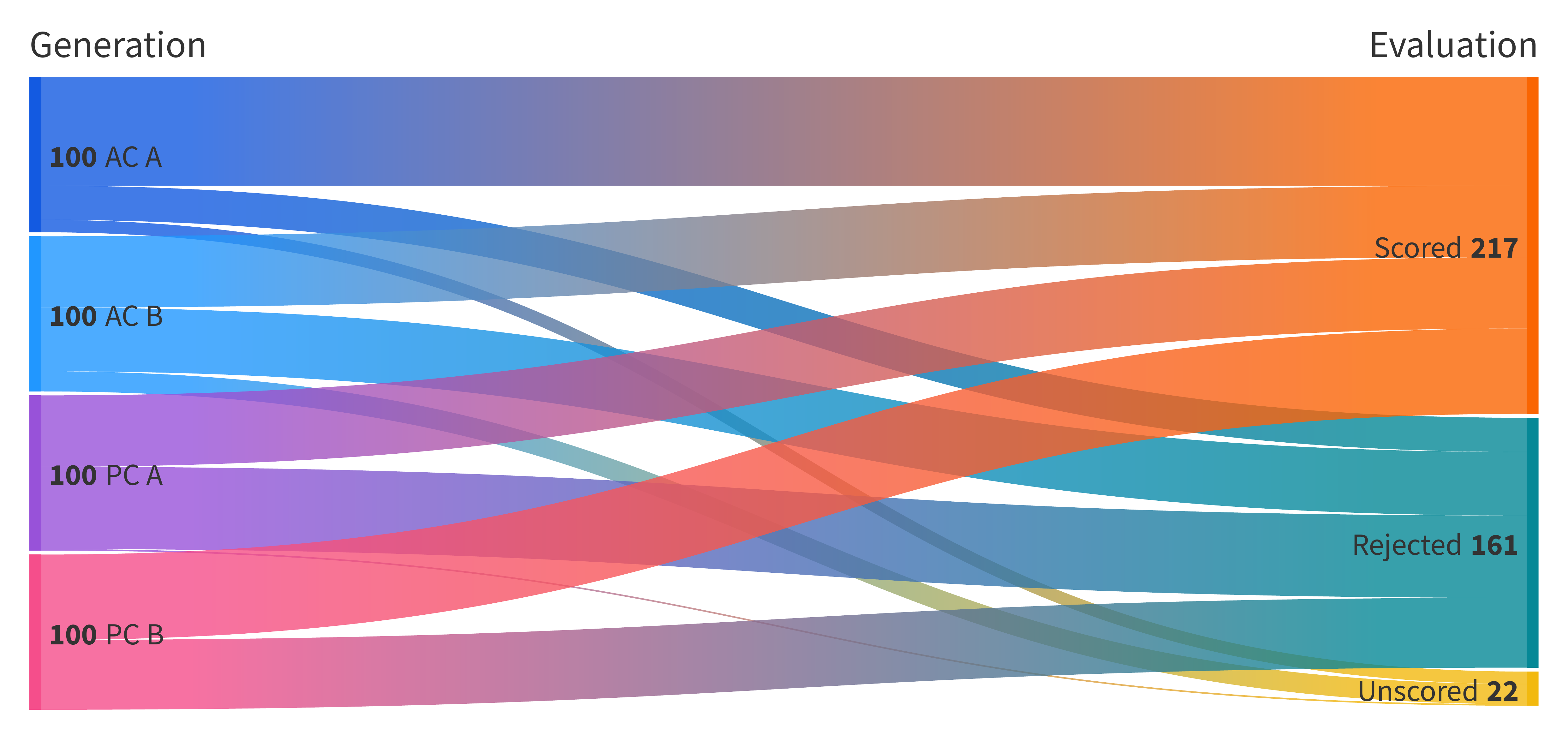}
\caption{Data flow from generation conditions to evaluation outcomes}
\label{fig:data_selection}
\end{figure}

Prior to detailed rating, a senior neuroradiologist performed an eligibility screening to identify synthetic images suitable for arterial-phase assessment. Following this step, 239 out of 400 generated images (59.8\%) were retained for subsequent multi-reader evaluation.
The retained set of 239 synthetic images was then assessed in detail by four blinded neuroradiologists using a structured Likert-scale protocol. For each image, raters independently evaluated the visibility and delineation of three arterial segments, including the proximal (large), medium, and peripheral vessels on a 5-point Likert scale (1 = very poor to 5 = excellent anatomical fidelity). In addition, raters could mark a segment as \textbf{not present} or \textbf{not applicable} if the corresponding vessels were genuinely absent from the field of view or if the image did not allow a meaningful judgment. A separate flag was provided to indicate \textbf{external circulation} where the depicted vasculature did not belong to the intended intracranial AC or PC territory, as shown in Table~\ref{tab:screening_overall}. Images were presented in random order, and raters were blinded to the conditioning labels and to each other's scores. The full evaluation flow is summarized in Fig.~\ref{fig:data_selection}.

We first aggregated scores at the image level. Segment-wise image scores were obtained by averaging the available Likert ratings across raters while ignoring entries marked as not present, not applicable, or external circulation. An image-level overall quality score was then computed as the mean of the proximal, medium, and peripheral segment scores that remained after this masking. Images for which all three segments were deemed non-ratable (i.e., all raters selected only exclusion flags) were excluded from the aggregated analysis (n=22, 5.5\%) and therefore do not contribute to Table~\ref{tab:imagewise_quality}. The per-condition number of images that entered this analysis, $N_{\text{images}}$, is reported in the third column of Table~\ref{tab:imagewise_quality}.

\begin{table}[htbp]
\centering
\resizebox{\textwidth}{!}{
\begin{tabular}{lcccccc}
\toprule
Circ.& Plane & $N_{\text{images}}$ & Overall & Prox. & Med. & Peri. \\
\midrule
AC & A & 70 & 3.34 $\pm$ 0.50 & 3.51 $\pm$ 0.66 & 3.29 $\pm$ 0.66 & 3.18 $\pm$ 0.74 \\
AC & B & 46 & 3.09 $\pm$ 0.72 & 3.36 $\pm$ 0.91 & 3.04 $\pm$ 0.87 & 3.00 $\pm$ 0.93 \\
PC & A & 46 & 2.73 $\pm$ 0.80 & 3.00 $\pm$ 0.85 & 2.81 $\pm$ 0.98 & 2.68 $\pm$ 0.92 \\
PC & B & 55 & 3.29 $\pm$ 0.50 & 3.54 $\pm$ 0.51 & 3.32 $\pm$ 0.54 & 3.07 $\pm$ 0.72 \\
\bottomrule
\end{tabular}}
\caption{Image-wise overall and segment-wise arterial quality across conditioning settings (Likert 1--5)}
\label{tab:imagewise_quality}
\end{table}

Table~\ref{tab:imagewise_quality} summarizes the resulting image-wise quality scores across the four conditioning settings. Overall arterial quality scores fall in the range of approximately 3.1--3.3 for three of the four conditions, corresponding to intermediate-to-good vessel visibility on the 5-point scale. The posterior circulation with Plane~A shows a noticeably lower mean overall score of 2.73~$\pm$~0.80, suggesting that this combination is more challenging for the model and yields images with lower and more variable perceived quality. Across all conditions, proximal vessels tend to receive slightly higher scores than medium and peripheral branches, while peripheral segments show the lowest mean scores, indicating that the model has more difficulty synthesizing fine distal vasculature than large proximal trunks.

\begin{figure}[htbp]
  \centering
  \begin{subfigure}[t]{\textwidth}
    \centering
    \includegraphics[width=\textwidth]{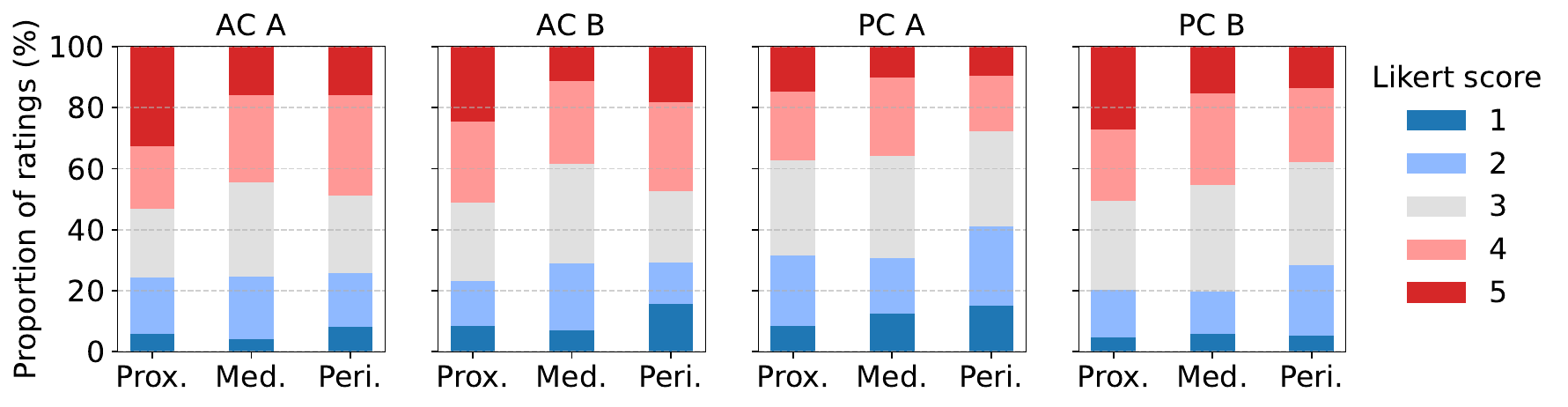}
    \caption{}
    \label{fig:likert_bar}
  \end{subfigure}
  \begin{subfigure}[t]{0.58\textwidth}
    \centering
    \includegraphics[width=\textwidth]{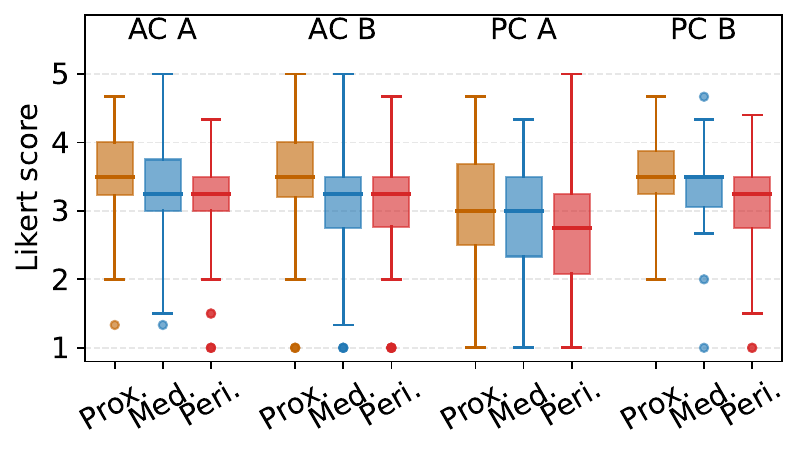}
    \caption{}
    \label{fig:likert_box}
  \end{subfigure}
  \begin{subfigure}[t]{0.4\textwidth}
    \centering
    \includegraphics[width=\textwidth]{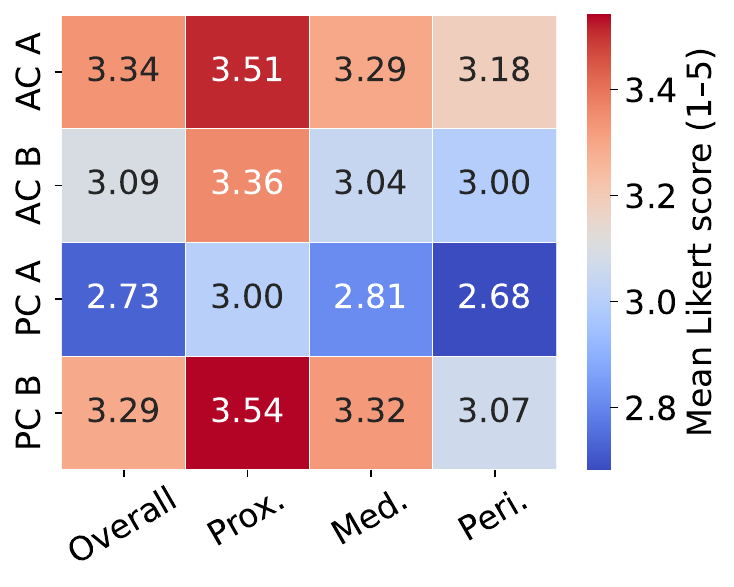}
    \caption{}
    \label{fig:likert_heatmap}
  \end{subfigure}
  \begin{subfigure}[t]{0.6\textwidth}
    \centering
    \includegraphics[width=\textwidth]{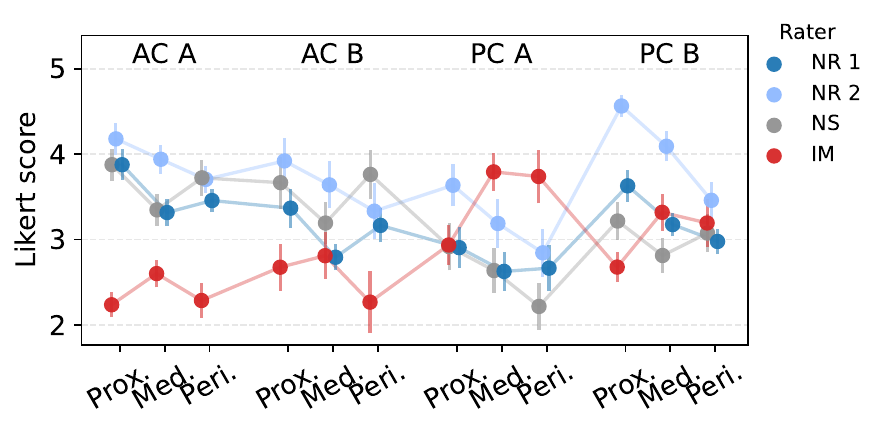}
    \caption{}
    \label{fig:likert_perReader}
  \end{subfigure}
  \begin{subfigure}[t]{0.38\textwidth}
    \centering
    \includegraphics[width=\textwidth]{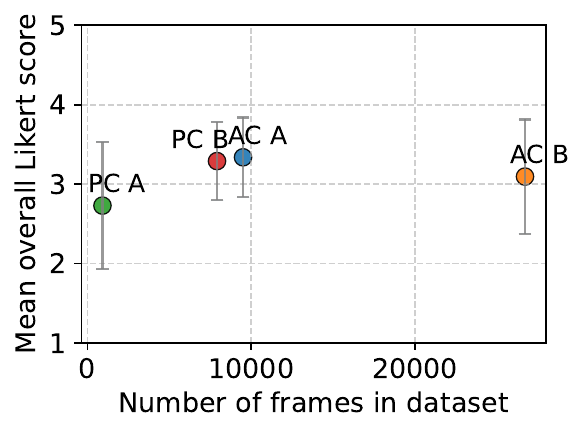}
    \caption{}
    \label{fig:likert_scatter}
  \end{subfigure}

  \caption{Reader study Likert ratings by conditioning setting and vessel segment. (a) Stacked proportional bar plots showing the distribution of raw 1–5 Likert ratings for each arterial segment (prox., med., peri.) across the four conditioning settings (AC/PC $\times$ Plane~A/B). (b) Boxplots of image-wise segment scores, obtained by averaging the available Likert ratings across raters for each image. Boxes indicate the interquartile range with median lines, whiskers extend to 1.5\,IQR, and points indicate outlying images beyond the whiskers. (c) Heatmap of the mean image-wise Likert scores across conditions and segments. Each cell shows the average rating (1–5). (d) Rater-wise mean Likert scores with 95\% confidence intervals. Raters include NR (neuroradiologist), NS (neurosurgeon), and IM (internal medicine expert). (e) Relationship between dataset richness and perceived image quality across conditioning settings}
\end{figure}


\paragraph{Distributional Analysis of Likert Ratings}
We characterized how radiologists distributed their ratings across the full Likert scale and across vessel segments by examining empirical score distributions rather than only their means. Individual ratings were summarized as per-condition proportional histograms for each arterial segment. The corresponding distributions are visualized in Fig.~\ref{fig:likert_bar}, highlighting how the proportion of high scores (4--5) varies across circulation, Plane, and vessel segment. To complement these raw-count summaries with an image-wise perspective, we then inspected the distribution of aggregated segment scores (one value per image and segment). Fig.~\ref{fig:likert_box} displays these image-wise scores as boxplots, making visible both central tendencies and variability across images within each condition. Furthermore, to provide a compact overview of how mean image-wise scores differ jointly across conditions and segments, we arranged the same summary statistics in a matrix layout (Fig.~\ref{fig:likert_heatmap}), which emphasizes relative strengths and weaknesses of the four conditioning settings. To further examine potential rater-specific effects, we also visualized mean Likert scores separately for each individual reader across vessel segments and conditioning settings, together with 95\% confidence intervals (Fig.~\ref{fig:likert_perReader}).

To further investigate why certain conditioning settings yield higher or lower perceived image quality, we examined the relationship between the prevalence of each condition in the real arterial-phase dataset and the mean overall Likert score assigned to the corresponding synthetic images. This analysis probes whether representation in the training distribution aligns with clinical quality in the generated outputs. As shown in Fig.~\ref{fig:likert_scatter}, no monotonic association emerges. Although AC B constitutes the most frequently observed configuration in the real dataset, it does not achieve the highest mean image quality. Conversely, PC A is underrepresented in the training data and exhibits markedly reduced synthetic quality, suggesting that extremely limited real-data coverage may constrain the model’s ability to synthesize diagnostically reliable vasculature for this condition. These findings highlight that dataset scale alone does not fully explain cross-condition variability and that anatomical–plane combinations differ in intrinsic generative difficulty.

Finally, we quantified the consistency of ratings across neuroradiologists using two-way random-effects intraclass correlation coefficients ICC(2,1) and ICC(2,\emph{k})~\cite{shrout1979intraclass}, treating images as targets and raters as a random effect. In this notation, the ``2'' denotes a two-way random-effects model with absolute agreement, the ``1'' index refers to the reliability of a single rater, and the ``\emph{k}'' index corresponds to the mean score across \emph{k} raters. These analyses were performed separately for the proximal, medium, and peripheral segments as well as for the overall image score derived from all three segments. The resulting coefficients, reported in Table~\ref{tab:icc_summary}, indicate moderate single-rater agreement and high reliability when averaging across the four raters. To assess the robustness of these estimates, we additionally report ICC values computed after restricting the analysis to domain experts only (two neuroradiologists and one neurosurgeon), which show consistently higher agreement.

\begin{table}[htbp]
\centering
\begin{tabular}{lcccc}
\toprule
\multirow{2}{*}{Region} 
& \multicolumn{2}{c}{All raters} 
& \multicolumn{2}{c}{Domain experts only} \\
\cmidrule(lr){2-3} \cmidrule(lr){4-5}
& ICC(2,1) $\uparrow$ & ICC(2,\emph{k}) $\uparrow$ & ICC(2,1) $\uparrow$ & ICC(2,k) $\uparrow$\\
\midrule
Proximal & 0.49 & 0.80 & 0.723 & 0.887\\
Medium & 0.56 & 0.84 & 0.731 & 0.891\\
Peripheral & 0.58 & 0.85 & 0.785 & 0.917\\
\midrule
Overall & 0.62 & 0.87 & 0.810 & 0.927 \\
\bottomrule
\end{tabular}
\caption{Inter-rater reliability (ICC(2,1) and ICC(2,\emph{k})) for overall and
segment-wise ratings, based on a two-way random-effects absolute-agreement
model~\cite{shrout1979intraclass}. Results are reported for (i) all raters and (ii) domain experts only. Here, domain experts refer to the two neuroradiologists and one neurosurgeon (\emph{k}=3), whereas the full rater set additionally includes one non-domain medical expert (\emph{k}=4). ICC(2,1) reflects the reliability of a single rater, whereas ICC(2,\emph{k}) reflects the reliability of the mean rating across \emph{k} raters}
\label{tab:icc_summary}
\end{table}

\subsection{Quantitative Evaluation}

To complement the reader study, we quantified the distributional similarity between synthetic and real DSA frames using the Fr\'echet Inception Distance (FID)~\cite{heusel2017gans}. FID compares the means and covariances of feature embeddings extracted from two image sets under a fixed encoder, and has become a standard summary statistic for assessing how closely a generative model matches the target data distribution.

For our evaluation, we sampled in total 10,000 arterial-phase frames from the conditional diffusion model, drawing 2,500 images for each of the four conditioning settings used in the reader study (AC/PC $\times$ Plane~A/B). As a reference set, we randomly selected 5,000 arterial-phase frames from the held-out validation data, approximately matching the same empirical distribution over the four conditions. All images were resized to $256 \times 256$ pixels, replicated to three channels, and passed through an ImageNet-pretrained Inception-V3 network to extract 2048-dimensional pool3 features. We then computed the FID between the real and synthetic feature distributions, both overall and separately by conditioning setting. The overall model achieved an FID of 15.27, indicating that the synthetic DSAs closely match real arterial-phase images in terms of global appearance characteristics such as intensity distributions, contrast patterns, and vascular texture.

\begin{table}[htbp]
  \centering
  \scriptsize
  \begin{tabular}{lc}
    \toprule
    \textbf{Publication} & \textbf{FID} $\downarrow$ \\
    \midrule
    DARL~\cite{kim2023Diffusion} & 177.59 \\
    HBGM~\cite{ovalle-magallanes2022Improving} & 84.09 \\
    MedNeRF~\cite{kshirsagar2024Generative} & 110.76 \\
    DDN~\cite{Liu2024} & 351.59 \\
    \midrule
    \textbf{Ours} & \textbf{15.27} \\
    \bottomrule
  \end{tabular}
  \caption{Reported FID scores in generative modeling studies for DSA or X-ray angiography. 
  Values are not directly comparable across works due to differences in task formulation (e.g., pseudo-angiogram generation for segmentation, native-to-DSA translation) and evaluation protocols}
  \label{tab:fid_scores}
\end{table}

It is important to interpret these values in the context of prior work. Table~\ref{tab:fid_scores} reports FID scores from recent DSA and X-ray angiography studies alongside our result. However, most existing methods do not perform synthesis of DSAs from noise: DARL~\cite{kim2023Diffusion} and HBGM~\cite{ovalle-magallanes2022Improving} generate pseudo-angiograms primarily to support downstream tasks such as segmentation or registration, while MedNeRF~\cite{kshirsagar2024Generative} and DDN~\cite{Liu2024} focus on view-synthesis or native-to-DSA image translation. In contrast, our model performs full generative synthesis from pure noise, with only high-level semantic conditioning (circulation, plane, angle) and no image-based guidance. Thus, while the numerical comparison is not a like-for-like benchmark, reporting prior FID scores provides context for the scale of values typically seen in angiography-related image synthesis.

Moreover, FID in this setting is computed using an encoder trained on natural images rather than angiography, and is sensitive to preprocessing choices such as image scaling and intensity normalization. We therefore view FID primarily as an internal consistency check rather than a standalone indicator of clinical usability. In combination with the reader study, the low FID score supports the conclusion that the diffusion model produces samples that are not only visually realistic to expert observers but also closely aligned with real DSAs in terms of global feature statistics.


\section{Discussion}

Our study demonstrates that semantically conditioned LDM can synthesize clinically realistic arterial-phase DSA frames under explicit control of anatomy and acquisition geometry. By curating a large arterial-phase DSA dataset with structured metadata, training a conditional diffusion model from noise, and validating the outputs through a multi-reader clinical study and quantitative distributional metrics, we showed that semantic conditioning enables controllable and high-fidelity cerebral DSA synthesis.

The reader study highlights systematic patterns in perceived image quality that align with established challenges in vascular imaging and prior work on synthetic angiography, CTA, and MRA. Across conditions, generated DSAs were generally rated as clinically plausible, while finer distal vasculature remained more difficult to reproduce than large proximal trunks. Although most prior neurovascular synthesis studies do not explicitly stratify image quality by arterial segment, reduced fidelity of peripheral vessels is frequently noted as a qualitative limitation in synthetic CTA and MRA~\cite{You2023,Koch2025}, reflecting the intrinsic sensitivity of small-caliber vessels to contrast, noise, and partial-volume effects. Our segment-wise reader evaluation provides a structured clinical confirmation of this widely recognized challenge in the context of synthetic DSA.

The relationship between data availability and generative quality is more nuanced than a simple monotonic dependence. Our scatter analysis relating the prevalence of each condition in the real arterial-phase dataset to the mean overall Likert score of the corresponding synthetic images (Fig.~\ref{fig:likert_scatter}) shows that the most frequent condition (AC~B) does not attain the best perceived quality, whereas the least represented condition (PC~A) stands out as clearly worse. This pattern suggests that extremely sparse coverage in the training data can limit the model’s ability to synthesize diagnostically reliable vasculature, but that beyond a certain threshold, other factors such as anatomical complexity, projection geometry, and conditioning design also govern generative difficulty. These observations emphasize the importance of balancing training sets across clinically relevant views, and of explicitly monitoring performance in underrepresented but important configurations.

Compared with existing angiography-related generative approaches, the present framework occupies a complementary point in the design space. Prior work has largely focused on task-specific formulations, such as generating pseudo-angiograms to support segmentation or registration~\cite{kim2023Diffusion,ovalle-magallanes2022Improving}, or on image-to-image mappings for view synthesis and native-to-DSA translation~\cite{gao2019Deep,kshirsagar2024Generative,Liu2024}. In contrast, our model performs full generative synthesis from pure noise in a latent space, using only high-level semantic conditioning on circulation, acquisition plane and C-arm angles, and is evaluated primarily in terms of standalone image realism. The resulting FID of 15.27 is substantially lower than previously reported values (Table~\ref{tab:fid_scores}), even though those scores were obtained for less demanding tasks that leverage an observed input image. While differences in evaluation protocol and feature extractor preclude a strict comparison, the combination of low FID and favourable reader-study outcomes indicates that semantically controlled diffusion models can produce virtual DSAs that are both quantitatively and clinically convincing.

Our study also highlights several limitations and directions for further work. First, the data originate from a single institution and a limited set of acquisition presets, and we only consider 2D frames rather than full dynamic sequences. Generalization to other centers, acquisition protocols, and temporally coherent sequence synthesis remains to be demonstrated. Second, conditioning is restricted to circulation and a small set of canonical C-arm angles; more fine-grained control over anatomy (e.g.\ specific vascular territories, stenoses, or aneurysms) and acquisition parameters could further increase the utility of the model for simulation and training. Third, quality assessment is based on expert visual ratings and an FID computed with an ImageNet-pretrained encoder; both are imperfect surrogates for clinical usefulness. Future work could incorporate domain-specific feature extractors, evaluate the impact of synthetic data on downstream tasks such as segmentation or detection, and probe whether the model ever hallucinates anatomically implausible or clinically misleading configurations.

Despite these limitations, our results demonstrate that diffusion models with explicit semantic conditioning on anatomy and C-arm geometry can generate large-scale libraries of realistic arterial-phase DSAs. Such synthetic angiograms may prove valuable as a privacy-preserving resource for method development, an augmentation source for data-hungry learning algorithms, and a tool for training or simulation in interventional neuroradiology. By combining high-fidelity generative modelling with rigorous clinical evaluation, this work takes a step towards integrating synthetic angiography into the broader ecosystem of data-driven cerebrovascular research.

\section{Methods}
\subsection{Study approval}
The clinical DSA data used in this study were acquired at the Department of Neuroradiology, Medical Faculty Mannheim, Heidelberg University and at the University Medical Center Mannheim (UMM). The study was approved by the Medical Ethics Commission II of Medical Faculty Mannheim, University of Heidelberg (approval nr.: 2017-825R-MA and 2017-828R-MA). The need for informed consent was waived due to the retrospective nature of the study. All procedures were in accordance with the ethical standards of the institutional and national research committee and with the 1964 Helsinki Declaration and its later amendments or comparable ethical standards.

\subsection{DSA Arterial-Phase Classification}

To facilitate the extraction of arterial-phase frames from raw DSA sequences, we trained a deep neural network to classify individual frames as either arterial-phase or non-arterial-phase. This classification task is formulated as a binary frame-level classification problem.

We began with a manually annotated subset of 953 DSA series, each labeled by clinical experts with a single representative arterial-phase frame. This curated subset was used to train a ResNet-18~\cite{he2016deep} classifier from scratch. All input frames were resized to $128 \times 128$ and normalized to the range $[0,1]$. Standard data augmentation techniques, including random cropping, horizontal flipping, and intensity jittering, were applied to enhance generalization.

The model maps each input DSA frame \( x \in \mathbb{R}^{H \times W} \) to a probability score \( \hat{y} \in [0,1] \) indicating whether the frame belongs to the arterial phase.
We train the network using the binary cross-entropy loss
\begin{equation}
\mathcal{L}_{\text{BCE}} = - \frac{1}{N} \sum_{i=1}^N \left[ y_i \log(\hat{y}_i) + (1 - y_i) \log(1 - \hat{y}_i) \right],
\end{equation}
where \( y_i \in \{0,1\} \) is the ground-truth label and \( \hat{y}_i \) is the predicted probability for the \( i \)-th frame from the $N$ available samples.

This automated classification process enabled large-scale identification of high-confidence arterial-phase frames without the need for exhaustive manual annotation. The resulting frame set served as input data for training our generative models.
After training, the classifier was applied to the remaining unannotated DSA series to identify arterial-phase frames. For each series, frames with the highest predicted arterial probabilities were selected as candidates.  Detailed training configuration and performance metrics are provided in the Experiment section.

\subsection{Unconditional Generation}

Our unconditional generative pipeline follows the Latent Diffusion Model (LDM) framework~\cite{rombach2022high}, where diffusion is applied in a learned latent space instead of the pixel space. This approach significantly reduces computational cost while maintaining high fidelity in the synthesized images.

\paragraph{Latent Autoencoder}
We begin by training a convolutional autoencoder to map each DSA frame \( x \in \mathbb{R}^{256 \times 256} \) into a lower-dimensional latent representation \( z = \mathcal{E}(x) \), where \( \mathcal{E} \) denotes the encoder. The decoder \( \mathcal{D} \) is used to reconstruct the original image from the latent space, \( \hat{x} = \mathcal{D}(z) \). The autoencoder is trained using a reconstruction loss
\begin{equation}
\mathcal{L}_{\text{recon}} = \mathbb{E}_{x \sim p_{\text{data}}} \left\| x - \mathcal{D}(\mathcal{E}(x)) \right\|_1,
\end{equation}
where $p_{\text{data}}$ denotes the empirical data distribution of arterial-phase DSA frames in the training set, and $\|\cdot\|_1$ denotes the element-wise $\ell_1$ norm.

\paragraph{Latent Diffusion Process}
Once trained, the encoder \( \mathcal{E} \) is used to obtain latent codes \( z \), on which the diffusion model operates. Following the DDPM formulation~\cite{ho2020denoising}, we define a forward diffusion process that adds Gaussian noise $\epsilon$ to the latent variables $z_t$ at steps $t=0,\ldots,T$,
\begin{equation}
z_t = \sqrt{\bar{\alpha}_t} z_0 + \sqrt{1 - \bar{\alpha}_t} \epsilon, \quad \epsilon \sim \mathcal{N}(0, I),
\end{equation}
where noise is added gradually according to a predefined noise schedule $\beta_t$, with $\alpha_t = 1 - \beta_t$ and $\bar{\alpha}_t = \prod_{s=1}^t \alpha_s$, and $\bar{\alpha}_t$ controls how much of the original signal remains at step $t$. 
A UNet-based denoising network \( \epsilon_\theta \) is trained to predict the noise component \( \epsilon \) given the noisy latent \( z_t \) and timestep \( t \). The training objective is the simplified noise prediction loss
\begin{equation}
\mathcal{L}_{\text{DDPM}} = \mathbb{E}_{z, \epsilon, t} \left\| \epsilon - \epsilon_\theta(z_t, t) \right\|^2.
\label{DDPM}
\end{equation}

\paragraph{Sampling}
At inference time, we start from Gaussian noise \( z_T \sim \mathcal{N}(0, I) \) in the latent space, and apply the learned denoising model \( \epsilon_\theta \) iteratively from \( t = T \) to \( t = 1 \). The latent at each step is updated as
\begin{equation}
z_{t-1} = \frac{1}{\sqrt{\alpha_t}} \left( z_t - \frac{1 - \alpha_t}{\sqrt{1 - \bar{\alpha}_t}} \epsilon_\theta(z_t, t) \right) + \sigma_t \cdot \eta, \quad \eta \sim \mathcal{N}(0, I),
\end{equation}
where $\sigma_t$ controls the amount of stochastic noise injected at each reverse diffusion step, ensuring sufficient randomness during sampling. This update follows the standard DDPM sampling formulation~\cite{ho2020denoising}.
After completing all denoising steps, we obtain the clean latent \( z_0 \), which is then decoded using the decoder
\begin{equation}
\hat{x} = \mathcal{D}(z_0).
\end{equation}
This sampling scheme enables us to generate realistic arterial-phase DSA frames in an unconditioned setting.

\subsection{Conditional Generation}
\label{sec_cond_gen}

\begin{figure}[htbp]
\centering
\includegraphics[width=0.85\textwidth]{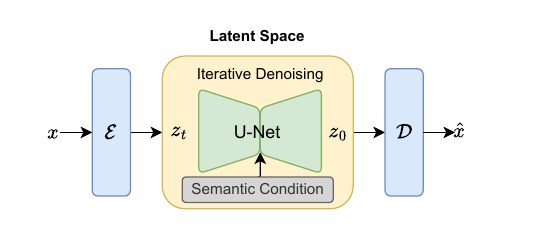}
\caption{Overview of the semantically conditioned latent diffusion framework for DSA synthesis}
\label{fig:ldmModel}
\end{figure}

Our conditional generation framework extends the latent diffusion setup by incorporating semantic metadata as conditioning information. Following~\cite{rombach2022high}, we train a BERT~\cite{devlin2019bert} encoder from scratch to embed textual descriptions of anatomical and acquisition-specific attributes associated with each DSA frame. Fig.~\ref{fig:ldmModel} provides an overview of the semantically conditioned latent diffusion pipeline in which iterative denoising is performed in latent space and guided by the semantic condition.

\paragraph{Metadata Construction}
For each image in the training dataset, we generate a descriptive sentence capturing three key attributes: the anatomical circulation type (anterior or posterior), the acquisition plane (e.g., Plane A or Plane B), and the primary and secondary angles.

\paragraph{BERT-based Embedding}
We train a BERT encoder to map each metadata sentence into a fixed-length embedding vector $c \in \mathbb{R}^d$. The encoder is trained jointly with the diffusion model and is not pre-trained on external corpora. The embedding $ c $ is projected to match the dimensionality expected by the denoising UNet and is injected into the model via cross-attention layers, allowing semantic alignment between condition and visual content.

\subsection{Implementation Details}

\paragraph{Latent Autoencoder (VAE)}

We implemented a convolutional variational autoencoder to map $256 \times 256$ DSA frames to a compact latent space with spatial resolution $64 \times 64$ and 3 channels. The encoder and decoder networks follow the configuration of the LDM framework~\cite{rombach2022high}, using 2 downsampling blocks with channel multipliers \{1, 2, 4\}, each containing two residual blocks. No attention layers are used. The VAE is trained using a perceptual reconstruction loss (LPIPS) combined with KL divergence:
\[
\mathcal{L}_{\text{VAE}} = \mathcal{L}_{\text{LPIPS}} + \lambda_{\text{KL}} \cdot D_{\text{KL}}(q(z|x) \parallel p(z)),
\]
where $q(z|x)$ denotes the variational posterior distribution produced by the encoder given an input image $x$, while $p(z)$ denotes the prior distribution over latent variables, which is chosen as a standard normal distribution $\mathcal{N}(0, I)$, and $\lambda_{\text{KL}} = 10^{-6}$. The input and output images are grayscale with a single channel. We use a batch size of 20 and train the model up to 100 epochs with the learning rate $2 \times 10^{-5}$, using early stopping with a patience of 10 epochs based on the validation loss.

\paragraph{Unconditional Latent Diffusion}
The unconditional LDM was trained using the latent codes produced by the VAE encoder described above. The model operates on latent tensors of size $3 \times 64 \times 64$, where diffusion is performed using a UNet-based denoising architecture. The diffusion process follows a linear noise schedule with 1000 steps, with $\beta_t$ linearly increasing from 0.0015 to 0.0195.
The denoising network is a UNet with 4 resolution levels and a channel configuration of $\{1, 2, 4, 4\}$ scaled by a base channel size of 224. Cross-resolution attention is applied at downsampling factors of $\{8, 4, 2\}$, corresponding to latent spatial resolutions of $\{8 \times 8, 16 \times 16, 32 \times 32\}$. Each block contains two residual layers, and multi-head attention uses 32 channels per head.
The model is trained to minimize the noise prediction objective defined in Eq.~\ref{DDPM}, where $\epsilon$ is standard Gaussian noise and $z_t$ is the noisy latent at time step $t$.
We used a batch size of 96, a learning rate of $5 \times 10^{-5}$, and trained the model for up to 500 epochs with early stopping.

\paragraph{Conditional Latent Diffusion}

We extend the unconditional LDM framework by introducing semantic conditioning using structured metadata. The conditioning information includes anatomical region (anterior or posterior), acquisition plane (e.g., Plane A, Plane B), and acquisition angles. Each image is paired with a descriptive sentence synthesized in a template-driven manner (see Sec.~\ref{sec_cond_gen} for examples). 
To embed these sentences, we employ a lightweight BERT encoder~\cite{devlin2019bert} with 4 transformer layers and an embedding size of 512. The encoder is trained jointly with the diffusion model from scratch. The resulting text embedding is injected into the UNet via cross-attention layers, enabling semantic control over the generative process.
The model architecture remains similar to the unconditional setting, using a 4-level UNet with channel multipliers $\{1, 2, 4, 4\}$ and base channel size of 224. Spatial attention is applied at multiple resolutions (8, 16, 32). The attention context dimension is set to 512 to match the BERT output. We use one transformer block per attention layer to balance expressivity and efficiency.
The training objective remains the same as the unconditional case (Eq.~\ref{DDPM}). However, the denoising network now takes both the noisy latent $z_t$ and the conditioning embedding $c$ as input:
\[
\mathcal{L}_{\text{cond}} = \mathbb{E}_{z, \epsilon, t, c} \left\| \epsilon - \epsilon_\theta(z_t, t, c) \right\|^2.
\]
We train the model with a batch size of 96 for up to 50 epochs using a learning rate of $5 \times 10^{-5}$.

\section{Data availability}
All synthetic DSA images generated by the proposed diffusion model and used in the reader study are available from the corresponding author upon reasonable request.



\clearpage
\printbibliography
\end{document}